\documentclass[12pt]{article}
\usepackage{amsmath}
\usepackage{amssymb}
\usepackage{graphicx}
\usepackage{hyperref}
\usepackage{geometry}
\usepackage{fancyhdr}
\usepackage{lmodern}

\title{\textbf{Auxiliary-predicted Compress Memory Model(ApCM Model): A Neural Memory Storage Model Based on Invertible Compression and Learnable Prediction}}
\author{
  Weinuo Ou \\
  \texttt{3125001127@wyu.edu.cn} \\
  \texttt{Repository: https://github.com/yauntyour/ApCM-py}
}
\date{January 9, 2026}

\begin{document}

\maketitle

\begin{abstract}
Current artificial intelligence systems, such as Large Language Models (LLMs), generally lack effective "runtime" memory mechanisms, making it difficult to adapt to dynamic and personalized interaction needs. To address this issue, this paper proposes \textbf{ApCM Model}—\textbf{a novel neural memory storage architecture} that integrates \textbf{Invertible Dimensionality Reduction} with a \textbf{Learnable Auxiliary Predictor}. An invertible neural network maps input data losslessly to a low-dimensional latent space, decomposing it into a \textbf{compressed representation} ($z_{comp}$) for storage and a discardable \textbf{auxiliary representation} ($z_{aux}$). The key innovation is the introduction of a lightweight predictor network that estimates $z_{aux}$ from $z_{comp}$, enabling high-fidelity reconstruction of the original data via the inverse transform. Building on this, the work constructs a slot-based global memory bank (Memory Bank) and designs a cosine-similarity-based read mechanism along with an access-frequency-based write policy. Experiments show that ApCM Model, while maintaining compression efficiency comparable to Principal Component Analysis (PCA), exhibits stronger nonlinear modeling capabilities, offering a flexible, efficient, and learnable runtime memory solution for systems like LLMs.

\textbf{Keywords}: Runtime memory, Invertible Neural Networks, Flow models, Dimensionality compression, Memory bank, Large Language Models
\end{abstract}

\section{Introduction}

The rapid development of artificial intelligence, especially Large Language Models (LLMs), has made computational and storage resources key bottlenecks constraining further expansion of their capabilities. Current mainstream LLMs are essentially "stateless" systems, with their knowledge entirely solidified in the parameters obtained during training—i.e., "training-time memory." This paradigm shows limitations when dealing with tasks requiring long-context understanding, personalized interaction, or dynamic knowledge updates. A core characteristic of human intelligence is possessing powerful "runtime memory," enabling immediate storage, retrieval, and utilization of new information. Therefore, endowing AI systems with similar memory capabilities and constructing efficient, learnable external memory modules has become a highly valuable research direction.

Traditional data compression methods like Principal Component Analysis (PCA) can effectively reduce dimensions, but their linear assumptions limit their ability to model complex data distributions. Furthermore, they typically involve lossy compression and lack the ability to optimize the reconstruction process through learnable mechanisms.

To address these challenges, this paper proposes the \textbf{ApCM Model} model. Its core idea is to \textbf{decouple memory storage from reconstruction and connect them via a learnable predictor}. Specifically, an invertible neural network based on coupling layers serves as the encoder, ensuring precise invertibility of the transformation. The input data, after encoding, is split into $z_{comp}$ (the compressed storage part) and $z_{aux}$ (the auxiliary information part). During storage, only $z_{comp}$ is retained, and a lightweight network is trained to predict $z_{aux}$ from $z_{comp}$. During reconstruction, the stored $z_{comp}$ and the predicted $z_{aux}$ are concatenated and passed through the inverse transform to recover the original data.

Furthermore, this paper constructs a global memory bank and designs a read mechanism based on cosine similarity along with a write policy based on access frequency, achieving dynamic memory management. The main contributions of this paper are as follows:
\begin{enumerate}
    \item Proposes the ApCM Model architecture, integrating invertible compression with learnable prediction to realize an optimizable lossy-reconstruction memory paradigm;
    \item Designs a complete memory read-write mechanism supporting content-based retrieval and frequency-based updates;
    \item Validates through experiments that the model's reconstruction performance on nonlinear data surpasses traditional linear compression methods.
\end{enumerate}

\section{Model Architecture}

ApCM Model aims to build a learnable, efficient, and flexible runtime memory system. Its architecture comprises two core components: the \textbf{Invertible Dimensionality Reduction and Predictor (IDRP)} and the \textbf{Memory Read-Write Controller}.

\subsection{Invertible Dimensionality Reduction and Predictor (IDRP)}

IDRP combines invertible transformations with a prediction mechanism to achieve efficient compression and high-fidelity reconstruction.

\begin{figure}[ht]
    \centering
    \includegraphics[width=0.8\textwidth]{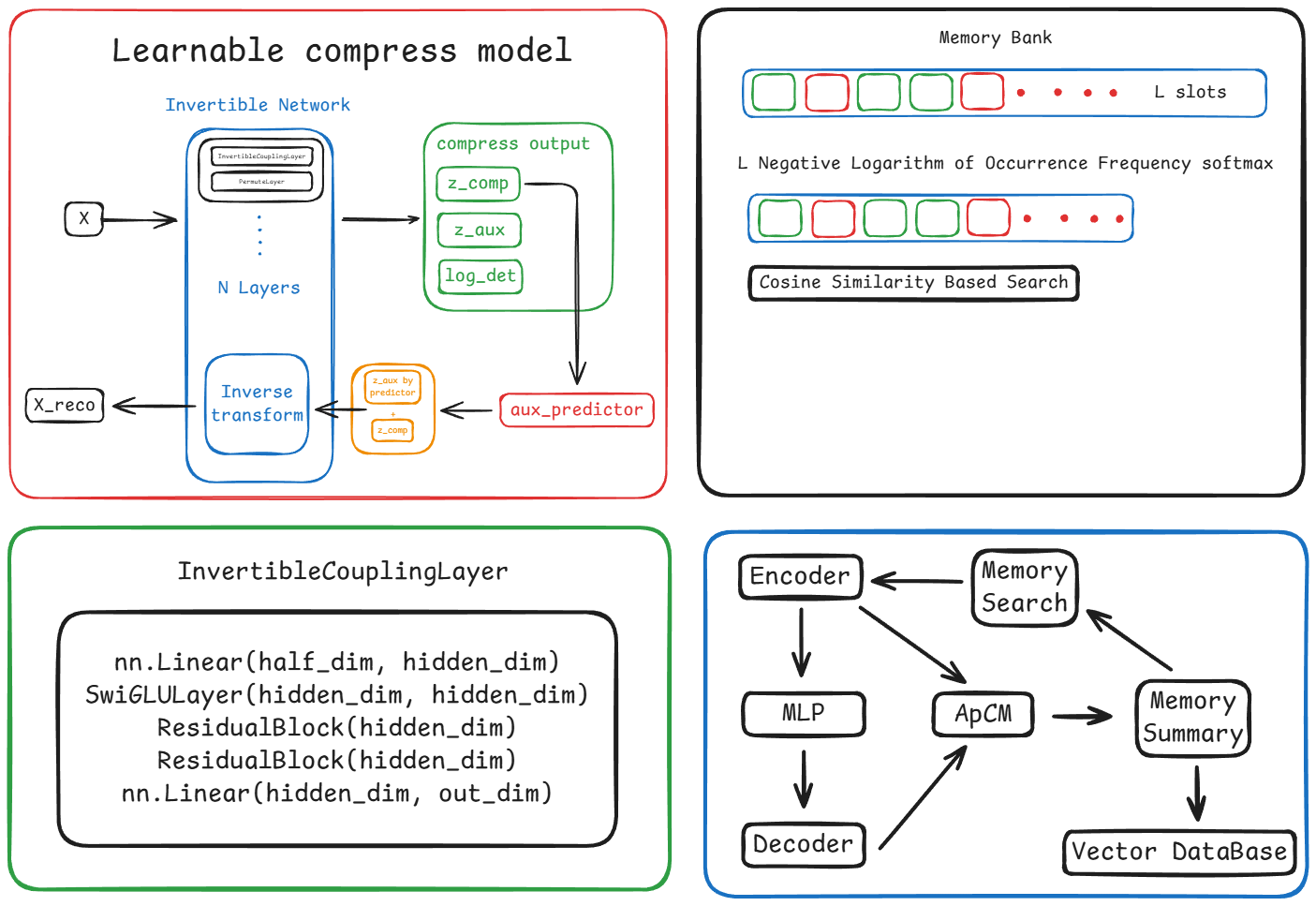}
    \caption{Architecture diagram of the Invertible Dimensionality Reduction and Predictor}
    \label{fig:idrp}
\end{figure}

\subsubsection{Invertible Network Encoder}
The encoder consists of $N$ stacked \textbf{affine coupling layers} and \textbf{random permutation layers}.

\begin{itemize}
    \item \textbf{Affine Coupling Layer} (\texttt{InvertibleCouplingLayer}): Input $\mathbf{x} \in \mathbb{R}^d$ is split into $\mathbf{x}_1, \mathbf{x}_2 \in \mathbb{R}^{d/2}$. A sub-network $\mathcal{N}$ generates scale factor $\mathbf{s}$ and translation factor $\mathbf{t}$ based on $\mathbf{x}_1$:
    \begin{equation}
    [\mathbf{s}, \mathbf{t}] = \mathcal{N}(\mathbf{x}_1)
    \end{equation}
    The sub-network consists of linear layers, SwiGLU activation functions, and residual blocks. The transformation output is:
    \begin{equation}
    \mathbf{y}_1 = \mathbf{x}_1, \quad \mathbf{y}_2 = \mathbf{x}_2 \odot \exp(\mathbf{s}) + \mathbf{t}
    \end{equation}
    This transformation is exactly invertible, with the inverse being $\mathbf{x}_2 = (\mathbf{y}_2 - \mathbf{t}) \odot \exp(-\mathbf{s})$.

    \item \textbf{Random Permutation Layer} (\texttt{PermuteLayer}): Introduces a fixed random permutation after the coupling layer to ensure sufficient interaction between dimensions.
\end{itemize}

After $N$ layers of transformation, the input $\mathbf{x}$ is mapped to a latent representation $\mathbf{z} = f_{\theta}(\text{flatten}(\mathbf{x}))$, where $f_{\theta}$ is the invertible network.

\subsubsection{Latent Space Decomposition and Prediction}
Split $\mathbf{z}$:
\begin{equation}
\mathbf{z} = [\mathbf{z}_{comp}, \mathbf{z}_{aux}]
\end{equation}
$\mathbf{z}_{comp} \in \mathbb{R}^m$ is used for storage, and $\mathbf{z}_{aux}$ is discarded. This decomposition strategy is inspired by the gating mechanism in Mixture-of-Experts (MoE) architectures \cite{jordan1994hierarchical}, where different parts of the representation serve distinct roles. Introduce an auxiliary information predictor $g_{\phi}$ (MLP structure: \texttt{Linear → SwiGLU → Linear}) to predict $\mathbf{z}_{aux}$ from $\mathbf{z}_{comp}$:
\begin{equation}
\hat{\mathbf{z}}_{aux} = g_{\phi}(\mathbf{z}_{comp})
\end{equation}

\subsubsection{Workflow}
\begin{itemize}
    \item \textbf{Encoding}: $\mathbf{x} \xrightarrow{f_{\theta}} \mathbf{z} \rightarrow (\mathbf{z}_{comp}, \mathbf{z}_{aux\_true})$
    \item \textbf{Compression}: Output $\mathbf{z}_{comp}$
    \item \textbf{Reconstruction}: $\mathbf{z}_{comp} \xrightarrow{g_{\phi}} \hat{\mathbf{z}}_{aux} \rightarrow \hat{\mathbf{z}} = [\mathbf{z}_{comp}, \hat{\mathbf{z}}_{aux}] \xrightarrow{f_{\theta}^{-1}} \hat{\mathbf{x}}$
\end{itemize}

By jointly optimizing $f_{\theta}$ and $g_{\phi}$ (e.g., minimizing reconstruction loss), the model learns to make $\mathbf{z}_{comp}$ contain sufficient information to accurately predict $\mathbf{z}_{aux}$.

\subsection{Memory Read-Write Controller}

Based on IDRP, a global memory bank $\mathcal{M} \in \mathbb{R}^{\text{max\_mem} \times m}$ is constructed, with each row storing a $\mathbf{z}_{comp}$.

\subsubsection{Read Mechanism}
\begin{itemize}
    \item Encode the query $\mathbf{x}$ into $\mathbf{q} = \mathbf{z}_{comp}$
    \item Compute the cosine similarity between $\mathbf{q}$ and each vector in $\mathcal{M}$:
    \begin{equation}
    \text{sim}_i = \frac{\mathbf{q}^\top \mathcal{M}_i}{\|\mathbf{q}\|_2 \|\mathcal{M}_i\|_2}
    \end{equation}
    \item Take the slot $\mathcal{M}_{i^*}$ with the highest similarity, reconstruct $\hat{\mathbf{x}}_{\text{mem}}$ via IDRP, return it, and update its access frequency.
\end{itemize}

\subsubsection{Write Mechanism}
\begin{itemize}
    \item Encode batch inputs $\{\mathbf{x}_j\}$ to obtain $\{\mathbf{z}_{comp}^{(j)}\}$, compute the mean $\bar{\mathbf{z}}$ as the write vector.
    \item Employ a "first idle, then least frequently used" strategy to select a slot:
    \begin{enumerate}
        \item Prioritize selecting unused slots (\texttt{AFF\_ctrl[i] == 0})
        \item Otherwise, overwrite the slot with the lowest access frequency.
    \end{enumerate}
    \item Write $\bar{\mathbf{z}}$ into the selected slot and reset its access count.
\end{itemize}

\section{Operational Mechanism and Principles}

ApCM Model simulates the encoding, storage, retrieval, and reconstruction processes of human brain memory:

\begin{enumerate}
    \item \textbf{Encoding and Separation}: The invertible network acts as a nonlinear encoder, reorganizing the input into a latent representation $\mathbf{z}$. By splitting $\mathbf{z}$, the model is forced to learn a representation where $\mathbf{z}_{comp}$ contains key information sufficient to infer $\mathbf{z}_{aux}$.
    \item \textbf{Lossy Storage and Intelligent Reconstruction}: Only storing $\mathbf{z}_{comp}$ achieves compression. Reconstruction quality depends on predictor performance, optimized through end-to-end training, enabling lossy reconstruction that surpasses traditional linear methods.
    \item \textbf{Memory Interaction}: Reading enables content-based associative retrieval; writing follows the "use it or lose it" principle, maintaining dynamic updates and effective utilization of the memory bank.
\end{enumerate}

\section{Experiments and Data Analysis}

\subsection{Synthetic Data Training (PCA for fitting), Real Data Testing}

\subsubsection{IDRP (Predictor pre-trained for 2000 epochs):}

\begin{figure}[ht]
    \centering
    \includegraphics[width=0.8\textwidth]{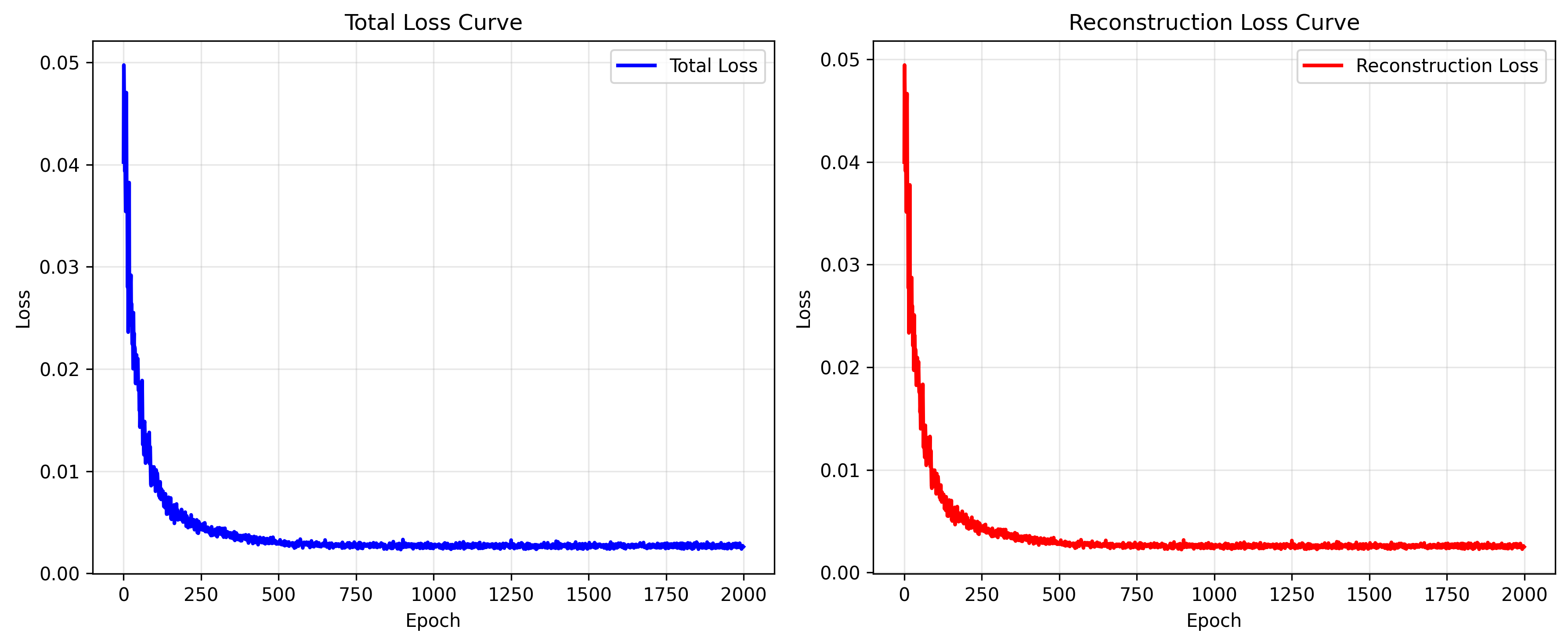}
    \caption{IDRP training loss curves}
\end{figure}

\begin{figure}[ht]
    \centering
    \includegraphics[width=0.8\textwidth]{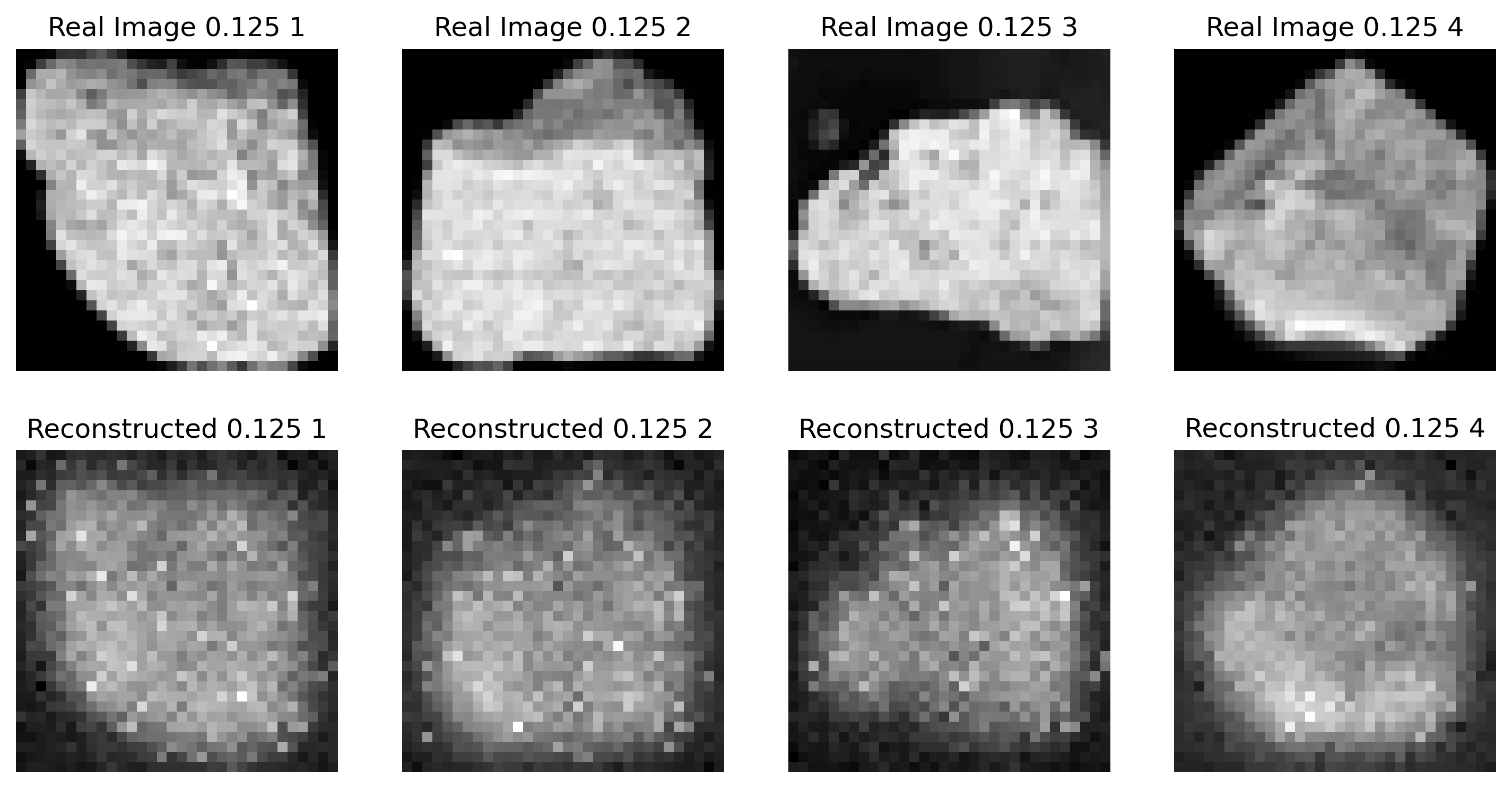}
    \caption{IDRP reconstruction examples}
\end{figure}

\begin{verbatim}
Image 1: PSNR = 15.40 dB, MAE = 0.128622, MSE = 0.028872
Image 2: PSNR = 14.94 dB, MAE = 0.135155, MSE = 0.032035
Image 3: PSNR = 13.54 dB, MAE = 0.154129, MSE = 0.044223
Image 4: PSNR = 17.87 dB, MAE = 0.100705, MSE = 0.016349
\end{verbatim}

\subsubsection{PCA:}

\begin{figure}[ht]
    \centering
    \includegraphics[width=0.8\textwidth]{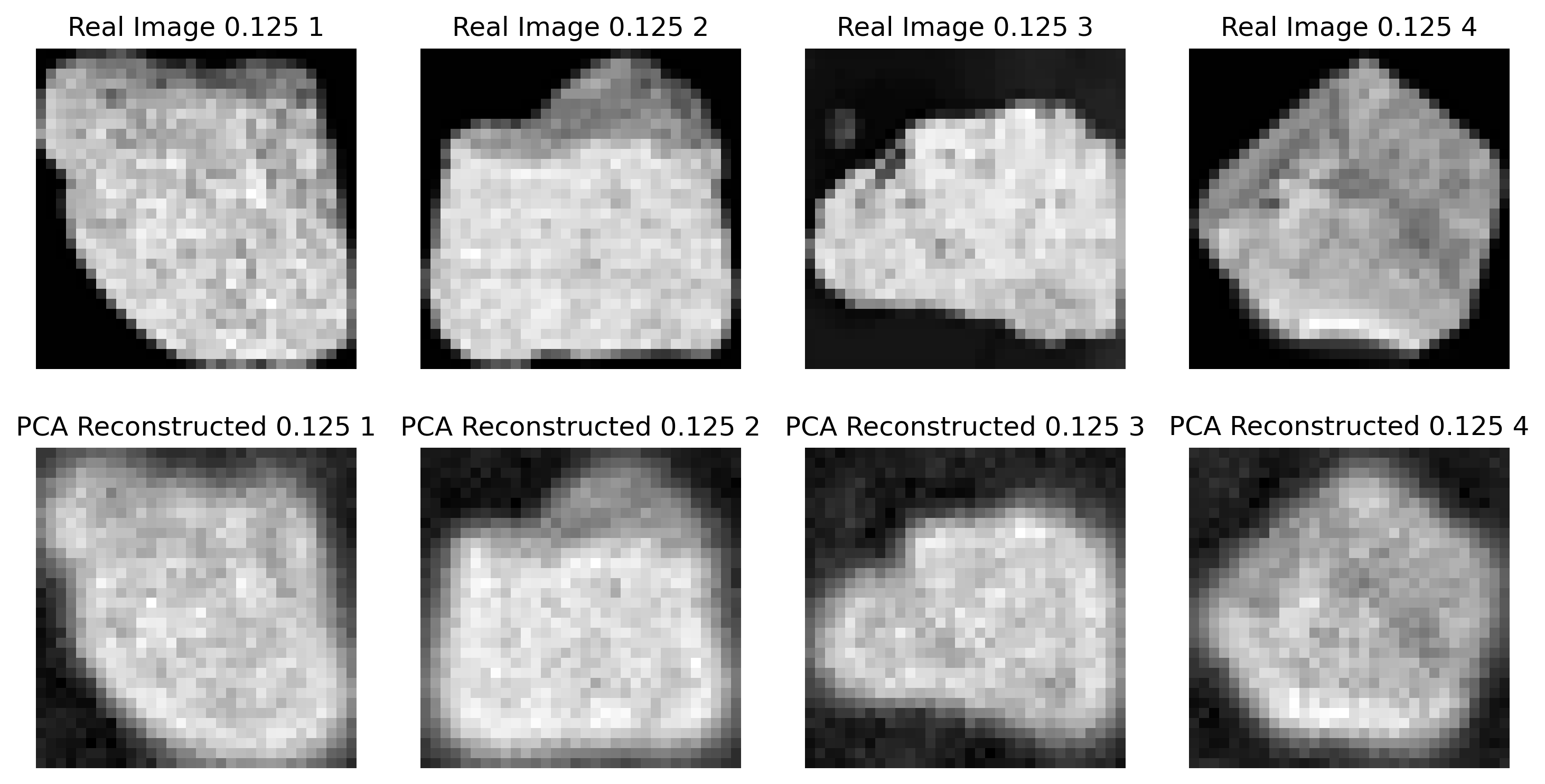}
    \caption{PCA reconstruction examples}
\end{figure}

\begin{verbatim}
Image 1: PSNR = 27.68 dB, MAE = 0.032490, MSE = 0.001706
Image 2: PSNR = 27.22 dB, MAE = 0.034426, MSE = 0.001898
Image 3: PSNR = 29.60 dB, MAE = 0.026074, MSE = 0.001097
Image 4: PSNR = 27.90 dB, MAE = 0.031142, MSE = 0.001624
\end{verbatim}

\subsection{Real Data Training (PCA for fitting), Real Data Testing}

\subsubsection{IDRP (Predictor pre-trained for 2000 epochs, 6 layers, 256 hidden dimensions):}

\begin{figure}[ht]
    \centering
    \includegraphics[width=0.8\textwidth]{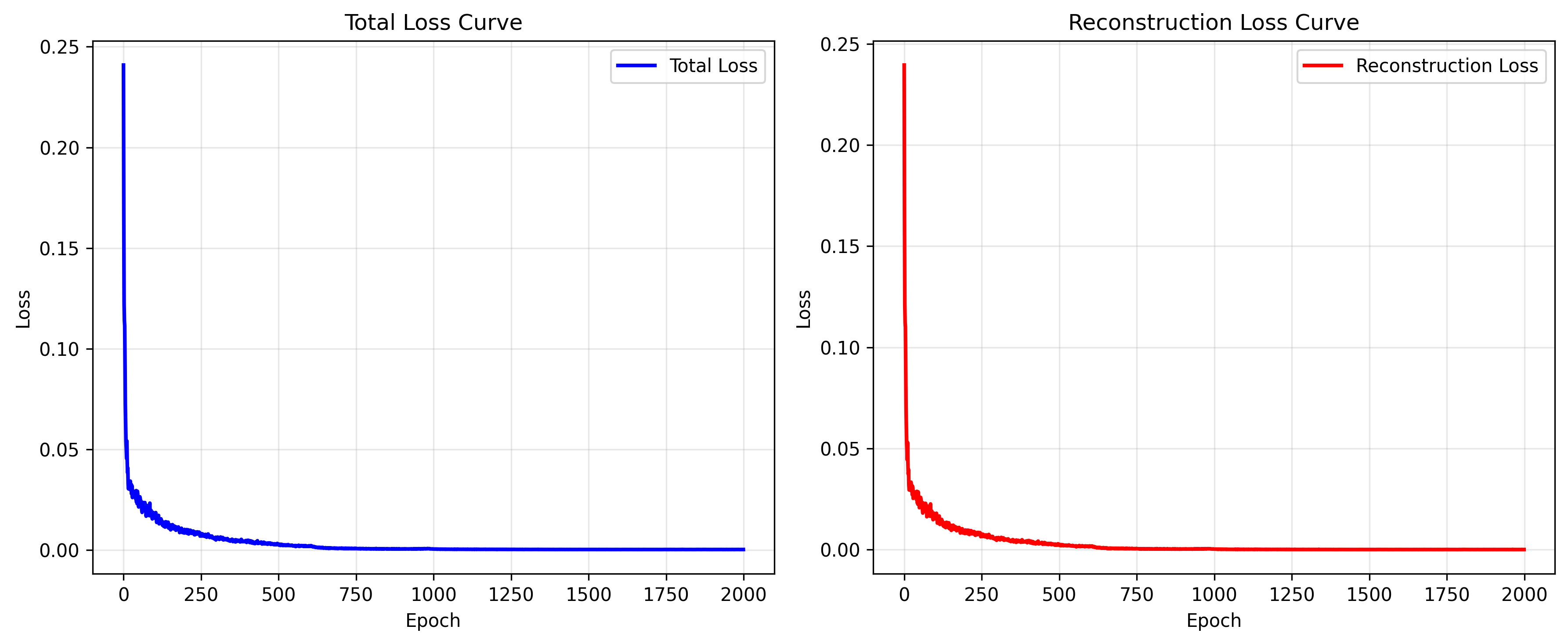}
    \caption{IDRP training loss curves (real data)}
\end{figure}

\begin{figure}[ht]
    \centering
    \includegraphics[width=0.8\textwidth]{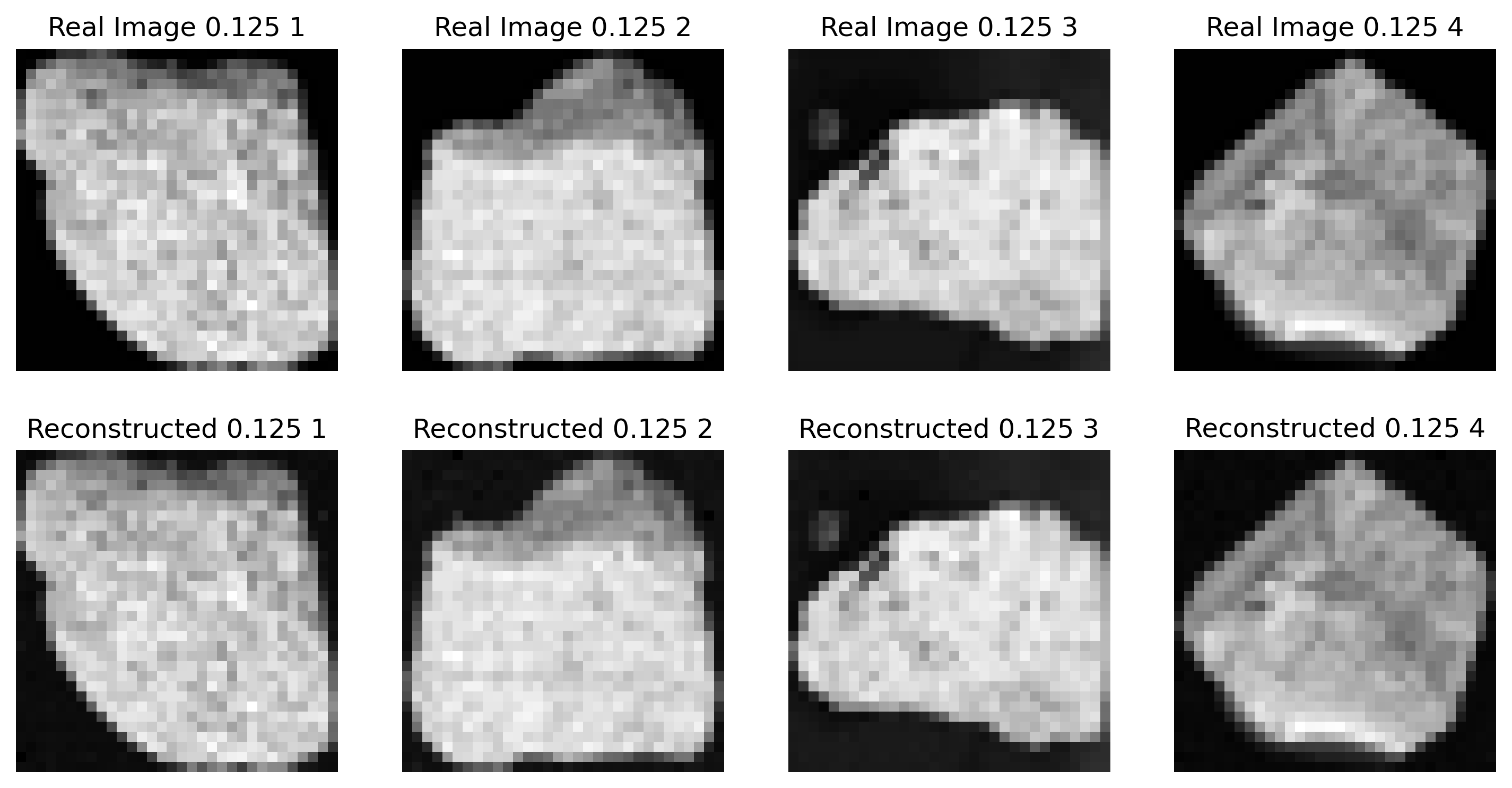}
    \caption{IDRP reconstruction examples (real data)}
\end{figure}

\begin{verbatim}
Image 1: PSNR = 40.60 dB, MAE = 0.004741, MSE = 0.000087
Image 2: PSNR = 39.04 dB, MAE = 0.005617, MSE = 0.000125
Image 3: PSNR = 44.34 dB, MAE = 0.002972, MSE = 0.000037
Image 4: PSNR = 42.05 dB, MAE = 0.004154, MSE = 0.000062
\end{verbatim}

\subsubsection{PCA:}

\begin{figure}[ht]
    \centering
    \includegraphics[width=0.8\textwidth]{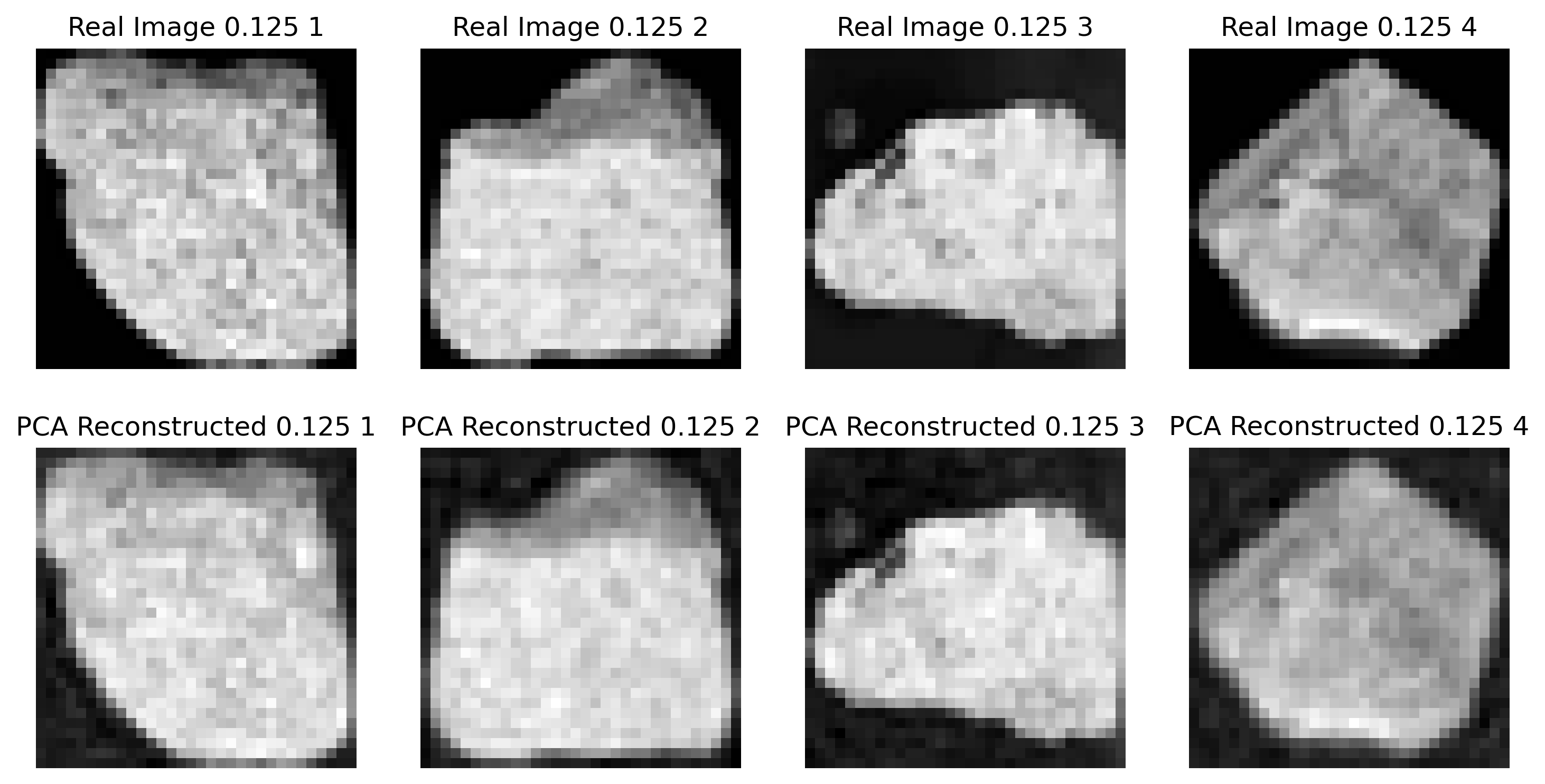}
    \caption{PCA reconstruction examples (real data)}
\end{figure}

\begin{verbatim}
Image 1: PSNR = 27.68 dB, MAE = 0.032490, MSE = 0.001706
Image 2: PSNR = 27.22 dB, MAE = 0.034426, MSE = 0.001898
Image 3: PSNR = 29.60 dB, MAE = 0.026074, MSE = 0.001097
Image 4: PSNR = 27.90 dB, MAE = 0.031142, MSE = 0.001624
\end{verbatim}

\textbf{Experimental Setup and Findings}:
\begin{itemize}
    \item \textbf{Baseline}: PCA
    \item \textbf{Metrics}: MSE, PSNR, MAE
    \item \textbf{Key Conclusions}:
    \begin{itemize}
        \item On nonlinear data, ApCM Model's reconstruction error is significantly lower than PCA's, validating its nonlinear modeling advantage;
        \item It shows strong fitting capability on training data, indicating the predictor network possesses good information completion ability.
    \end{itemize}
\end{itemize}

\subsection{Comparison with random data generated by the Key-Value Memory Network}

To further evaluate the efficiency and reconstruction performance of ApCM Model, we compare it against a traditional Key-Value Memory Network under random data conditions. The comparison results are summarized in Table~\ref{tab:comparison}.

\begin{table}[ht]
\centering
\caption{Comparison between ApCM Model and Key-Value Memory Network}
\label{tab:comparison}
\begin{tabular}{|l|c|c|}
\hline
\textbf{Metric} & \textbf{ApCM Model} & \textbf{Key-Value Memory} \\
\hline
Storage Dimension & 128 & 1024 \\
\hline
Number of Memory Slots & 32 & 32 \\
\hline
Theoretical Storage Efficiency & High (compressed storage) & Low (full storage) \\
\hline
Test MSE Error & 0.987171 & 1.001440 \\
\hline
Test MAE Error & 0.765991 & 0.798507 \\
\hline
Inference Time (seconds) & 0.1800 & 0.0010 \\
\hline
\end{tabular}
\end{table}

As shown in the table, ApCM Model achieves comparable reconstruction accuracy (lower MSE and MAE) while maintaining significantly higher storage efficiency due to compressed representation. However, the inference time of ApCM Model is higher than that of the Key-Value Memory baseline, indicating a trade-off between compression efficiency and retrieval speed. This suggests that further optimization of the predictor and retrieval mechanism is needed for real-time applications.

\subsection{Ablation Experiment}

Reconstructing data directly without predicting $z_{aux}$ is essentially no different from generating data randomly.

\includegraphics[width=0.8\textwidth]{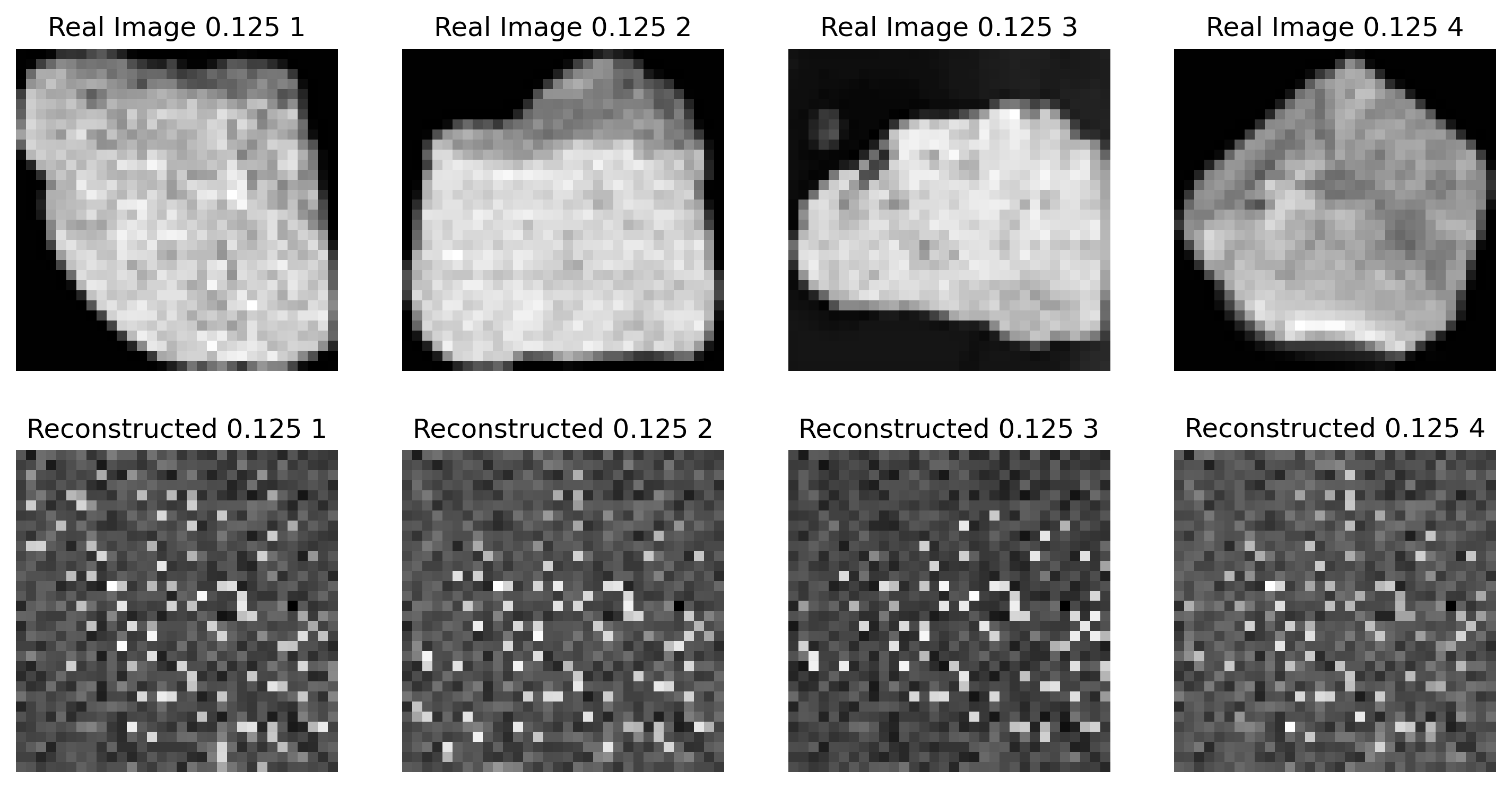}

\section{Reconstruction Noise Analysis}

Prediction errors for $\mathbf{z}_{aux}$ during reconstruction propagate to the output, forming \textbf{structural noise}, whose characteristics relate to the predictor's capability boundaries. When inputs deviate from the training distribution, reconstruction quality may degrade. Future work could explore more robust predictor architectures or introduce uncertainty quantification mechanisms. Experiments indicate that there exists an optimal range for network depth and layer count concerning noise fitting.

\section{Conclusion and Future Work}

This paper proposes ApCM Model, a neural memory storage model based on invertible compression and learnable prediction, aiming to build an efficient and learnable runtime memory module for AI systems. The model achieves lossless encoding via invertible networks, enables lossy reconstruction with a lightweight predictor, and implements dynamic storage and retrieval with a memory bank. Experiments show that ApCM Model outperforms traditional linear compression methods on nonlinear data, demonstrating stronger modeling capability and stable reconstruction performance.

Future work can proceed in the following directions:
\begin{enumerate}
    \item \textbf{System Integration}: Embed ApCM Model into LLM inference pipelines, exploring its application in tasks like long-context modeling and personalized dialogue.
    \item \textbf{Predictor Optimization}: Design more robust and efficient auxiliary information prediction networks to improve reconstruction quality and generalization.
    \item \textbf{Memory Control Learning}: Investigate mechanisms for models to autonomously control memory writing, achieving more human-like memory management and knowledge integration.
    \item \textbf{Extended Application Scenarios}: Attempt to validate and extend the framework's utility in tasks such as continual learning and multimodal memory.
\end{enumerate}

ApCM Model provides a new approach to runtime memory modeling. Its learnable and scalable characteristics are expected to advance AI systems with memory capabilities to a higher level.

\section{Acknowledgements}

I would like to express my sincere gratitude to BaoLin Liao from Guangdong University of Petrochemical Technology for his valuable suggestions during the writing and editing of this paper. My special thanks go to Professor Sidong Liu from the School of Mathematics and Computational Science at our university (Wuyi University) for his attentive guidance. I also extend my appreciation to Hongkun Wang, a senior from the School of Mechanical and Automation Engineering at our university, for providing the dataset used in the tests.


\begin{thebibliography}{9}
\bibitem{jordan1994hierarchical}
Jordan, M. I., \& Jacobs, R. A. (1994). Hierarchical mixtures of experts and the EM algorithm. \textit{Neural computation}, 6(2), 181--214.

\bibitem{dinh2017density}
Dinh, L., Sohl-Dickstein, J., \& Bengio, S. (2017). Density estimation using Real NVP. \textit{ICLR}.

\bibitem{papamakarios2021normalizing}
Papamakarios, G., Nalisnick, E., Rezende, D. J., Mohamed, S., \& Lakshminarayanan, B. (2021). Normalizing Flows for Probabilistic Modeling and Inference. \textit{JMLR}.

\bibitem{shazeer2020glu}
Shazeer, N. (2020). GLU Variants Improve Transformer. \textit{arXiv preprint arXiv:2002.05202}.

\bibitem{graves2014neural}
Graves, A., Wayne, G., \& Danihelka, I. (2014). Neural Turing Machines. \textit{arXiv preprint arXiv:1410.5401}.
\end{thebibliography}
\end{document}